\documentclass[letterpaper]{article} % DO NOT CHANGE THIS
\usepackage{aaai25}  % DO NOT CHANGE THIS
\usepackage{times}  % DO NOT CHANGE THIS
\usepackage{helvet}  % DO NOT CHANGE THIS
\usepackage{courier}  % DO NOT CHANGE THIS
\usepackage[hyphens]{url}  % DO NOT CHANGE THIS
\usepackage{graphicx} % DO NOT CHANGE THIS
\usepackage{natbib}  % DO NOT CHANGE THIS AND DO NOT ADD ANY OPTIONS TO IT
\usepackage{caption} % DO NOT CHANGE THIS AND DO NOT ADD ANY OPTIONS TO IT
\usepackage{algorithm}
\usepackage{algorithmic}

\usepackage{xcolor}
\usepackage{amsmath}
\usepackage{amssymb}
\usepackage{slashed}
\usepackage{subfigure}
\usepackage{multicol}

\usepackage{newfloat}
\usepackage{listings}
\DeclareCaptionStyle{ruled}{labelfont=normalfont,labelsep=colon,strut=off} % DO NOT CHANGE THIS
\floatstyle{ruled}
\newfloat{listing}{tb}{lst}{}
\floatname{listing}{Listing}
\title{Constructing sensible baselines for Integrated Gradients}
\author{
    Jai Bardhan\thanks{jai.bardhan90@gmail.com},
    Cyrin Neeraj\thanks{cyrin.neeraj@research.iiit.ac.in},
    Mihir Rawat\thanks{mihir.r@research.iiit.ac.in}, and
    Subhadip Mitra\thanks{subhadip.mitra@iiit.ac.in}
}
\affiliations{
    Center for Computational Natural Sciences and Bioinformatics, \\ 
    International Institute of Information Technology, Hyderabad 500 032, India.\\
}

\usepackage{bibentry}
\begin{document}

\maketitle

\begin{abstract}
  Machine learning methods have seen a meteoric rise in their applications in the scientific community. However, little effort has been put into understanding these ``black box'' models. We show how one can apply integrated gradients (IGs) to understand these models by designing different baselines, by taking an example case study in particle physics. We find that the zero-vector baseline does not provide good feature attributions and that an averaged baseline sampled from the background events provides consistently more reasonable attributions.  
\end{abstract}

\section{Introduction}

Machine learning (ML) models are now used in almost all branches of science, with various large-scale efforts like AlphaFold~\cite{Jumper2021} showing good promise. This has been possible mainly because of the current availability of large amounts of data and compute to train these models. As these models (and their impacts) grow and become more autonomous, it becomes crucial to question their \emph{understanding} of the world and gauge what influences their decisions. 

With this motivation, we explore the interpretability of the simplest deep learning architecture -- the fully connected deep neural network (DNN) used in many domains of science and engineering. For example, in the case of bioinformatics, morphological features from cell imaging, such as size, shape, texture of nuclei, etc., may be used as input to train the model to predict whether the cell is cancerous or not~\cite{cancers12030603}. Similarly, in the domain of astronomy, features derived from the light curves, such as rise time, peak luminosity, decay rate, etc., can be used as features for DNNs to classify supernovae from the background~\cite{Smith_2023}. 
In the case of High Energy Physics, features from particle collisions, such as the collision energy, jet types, etc., are used for predicting whether the collision has emerged from an interesting signal event~\cite{ATLAS:2019vwv}.

Approaches to interpretability can be broadly divided into two types. The first of the two types work by designing interpretable model architectures and training paradigms. These paradigms trade off performance for interpretability and are harder to work with, which has prevented their widespread adoption. The second type of interpretability works in a post-hoc approach where we try to understand the inner workings of the model which we have already trained. 

Here, we focus on post-hoc interpretability, in particular on a tool called Integrated Gradients (IGs)~\cite{10.5555/3305890.3306024}. IGs provide attributions for each input feature to indicate how relevant it was for the model's prediction for a particular sample. 

IG-based feature attributions are calculated by linearly interpolating along the feature space from the baseline to the data sample and evaluating the model prediction at each interpolated step. The choice of baseline depends on the problem at hand. For example, for image classification tasks, a completely dark image (zero pixel intensity) can be used as a baseline as each pixel is then devoid of any information; any image data sample can be constructed from a dark image to see which pixels are more important. An in-depth analysis of IGs in the context of image classification is found in Ref.~\cite{sturmfels2020visualizing}. This intuition of a baseline with zero information can be extended to classification tasks in particle physics. If jets are represented as images centred around the jet axis, such a baseline could be used to derive feature attributions. Ref.~\cite{Apolinario:2021olp} used randomly initialised images as baselines, which are then interpolated to correctly classify samples of quenched jets to derive pixel attributions. However, if we move away from the domain of image data to other data representations, there are no clear-cut intuitive prescriptions. For example, in jet classification using point clouds or event classification tasks where the data is represented as graphs or $n$-dimensional vectors of kinematic features. To use powerful tools like IGs, it is important to investigate baselines for each representation used in particle physics to understand the predictions of deep learning models. 

In this article, we investigate baseline constructions for a case study on event classification in particle physics, where events are represented by kinematic features. We train a model to distinguish a new physics signal (e.g., from heavy (vectorlike) quarks production) from the Standard Model backgrounds. The baselines will focus on the question that is of most interest to the particle physics searches: \emph{according to the DNN model, what makes the signal events differ from the background ones?} As typical new physics signal events at collider experiments occur at a tiny rate compared to those from backgrounds, this will help us to identify features that are important to tag signal events, compare the features' attributions to our intuitions to a reasonable degree and build better, economical models that will speed up the searches at the collider experiments.

\section{Integrated Gradients}
IGs explain an ML model's predictions by assigning an attribution score to each input feature, which reflects the contribution of that feature to the predictions. These attributions serve two main purposes: (1) they allow us to assess the trustworthiness of the model's prediction, and (2) they help us identify relevant and distinguishing features between the classes.

IGs work on the principle that the contribution of each input feature to a model's prediction can be estimated by integrating the gradient along a path from a baseline input to the actual input. Formally, the importance of the $i^{\text{th}}$ input is expressed as 
\begin{equation}
    \phi_i^{IG}(f, \mathbf{x}, \mathbf{x'}) = (x_i - x_i') \int_{\alpha = 0}^{1} \frac{\delta f(x' + \alpha (x - x'))}{\delta x_i} d\alpha,
    \label{eq:IG-mastereq}
\end{equation}
where $x$ and $x'$ represent the features of the input and the baseline input, respectively. 
The attribution score is blind to the common information between the baseline and the input $x$. 
Depending on the construction of $x'$, the attributions obtained from Eq.~\ref{eq:IG-mastereq} will be different and, therefore, the inferences from the model. 
However, we have a few additional domain-specific constraints: (1) the baseline must be from the data distribution, and (2) the baseline must be physically intuitive. Then, one would be able to infer from the feature attributions that, for example,
\emph{compared to the baseline, the features $f_1, f_2, \dots, f_k$ (in the order of importance) make this sample more $<$input class$>$-like as per the DNN model.}

If the baseline does not contain any information, one could claim, for example, 
\emph{the information in features $f_1, f_2, \dots, f_k$ (in the order of importance) when presented to the DNN makes the sample more $<$input class$>$-like.}
A baseline analogous to the dark image baseline in image classification tasks (i.e., a vector of all zeros) is unnatural for model inferences in particle physics.  
Firstly, the zero vector may not belong to the data distribution, i.e., there could be features that cannot even take the value $0$. Another reason is that the zero vector may not actually be a baseline void of any information; for instance, rotation angles in physics can take values $-\pi$ to $\pi$ where a value of $0$ would indicate something specific of the setup.  Thirdly, the zero baseline generally does not correspond to a physical object, which would make interpretations even more difficult.

\subsection{Averaged Baselines}

Averaging over the distribution of baselines can address the limitations of a single baseline, especially when it is unclear what the best baseline should be, i.e., when the baseline may lead to biased or extreme feature attributions. 
Instead of relying on a single baseline, multiple baselines are sampled from a distribution $D$, and integrated gradients are calculated from each baseline and averaged to give the final feature attributions. Mathematically, it is expressed as
\begin{equation}\label{eq:avg-ig}
    \phi_i(f, \mathbf{x}, \mathbf{x'}) = \int_{\mathbf{x'}} \phi_i^{IG}(f, \mathbf{x}, \mathbf{x'}) \times p_D(\mathbf{x'}) d\mathbf{x'}, 
\end{equation}
where $x'$ is now sampled from a distribution $D$, and $p_D$ symbolizes the probability density function.

Such a construction is especially suited for event classification problems in particle physics. In colliders, the detectors can detect only known particles. In the signal events of new particle searches, the detected particles will originate in some New Physics processes, whereas background events come from Standard Model processes. These processes are kinematically distinct (e.g., different mass/energy scales). The signal events must be separated from the background ones, which occur at much higher rates. A baseline constructed over the background events helps us investigate how the model distinguishes the signal events from the background ones.

\section{Our baselines for event classification}

The data representation of interest in this paper is $n$-dimensional kinematic data, which we will discuss in detail in the context of the heavy quark signal classification. We construct the following baselines to study feature attributions.
\medskip

\noindent \textbf{Averaging over background events as baseline:} For a binary classification problem such as ours, the background class does not contain any information about the signal class. Hence, one can construct baselines on a particular event from the background classes to probe model predictions on what makes the signal events different from the background one considered. 
The inference, then, becomes:
% \begin{center}
\emph{with respect to this specific background event, features $f_1, f_2, \dots, f_k$ (in the order of importance) of input event makes it more signal-like as per the DNN model.}
However, we might still miss out on attributions for features for which the baseline has similarities with the input; for example, a baseline with a particular high $H_T$ would give low attributions to $H_T$.
% \end{center}
\begin{figure*}[t!]
    \begin{center}
    \subfigure[Top-$5$ Features, B$_0$]{\includegraphics[width=0.31\textwidth]{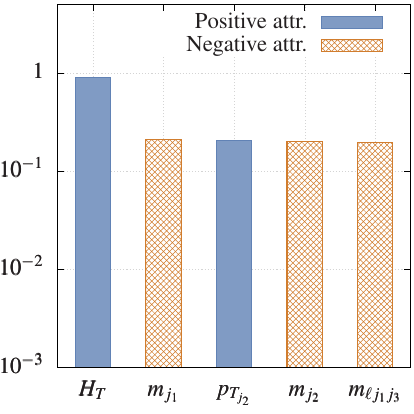}}\hfill
    \subfigure[Top-$5$ Features, B$_{\text{bg}}$]{\includegraphics[width=0.31\textwidth]{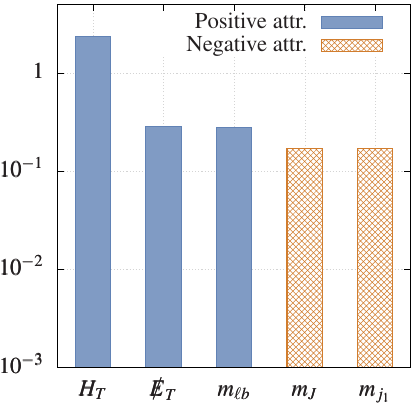}}\hfill
    \subfigure[Top-$5$ Features, B$_{\text{bgw}}$]{\includegraphics[width=0.31\textwidth]{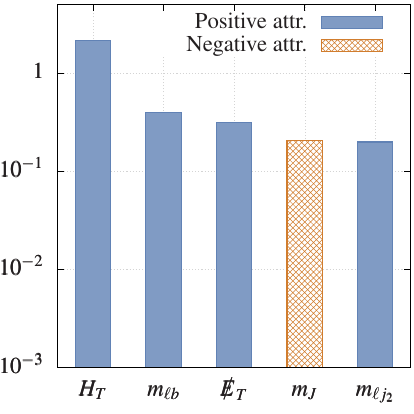}}\hfill
    \caption{Top-$5$ features ranked by their attribution scores for each of the baselines. }
    \label{fig:top5}
    \end{center}
% \vskip -0.5em
\end{figure*}
Therefore, we propose to construct the baseline by averaging over all the background events (in a particular search), i.e., we set $D$ to be the background distribution in Eq.~\ref{eq:avg-ig}. This choice is well motivated from the physics perspective as we want to identify features relevant to the signal when compared to the background and can effectively distinguish the signal from the background events. Unlike most other domains, our background class is actually composed of multiple background processes with different cross-sections. 

This setup enables us to examine various weighting schemes for averaging within the background processes -- namely, an unbiased weighting scheme where all the background processes receive equal weight (i.e., $p_{D_{bg}} \equiv \mathcal{U}$, where $\mathcal{U}$ is uniform distribution), and a weighting scheme proportional to the cross-sections of the processes (i.e., $p_{D_{bg}} \equiv S $, where $S$ is the density function proportional to the cross-sections of the background process). Attributions from the first baseline represent features relevant for distinguishing from an unbiased average of the backgrounds. In contrast, the second represents features relevant for distinguishing from the true (at LHC) weighted average of background. 
\medskip

\noindent \textbf{Blank baseline:} We also consider a naive choice for baseline, i.e., blank (zero-vector) baseline. This baseline is the standard choice for interpreting image classification models. As mentioned earlier, we expect this to perform poorly compared to the averaged baseline.

\begin{figure*}
    \begin{center}
    \subfigure[Accuracy]{\includegraphics[width=0.44\textwidth]{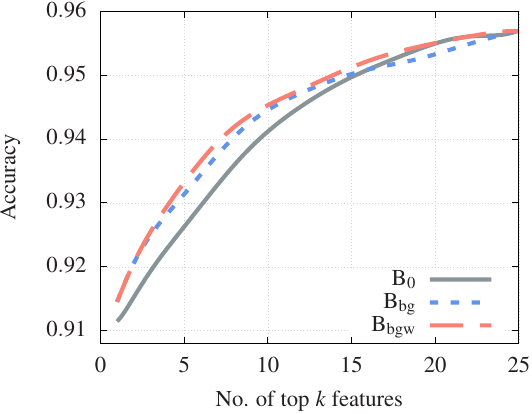}}\hfill
    \subfigure[$Z$-score]{\includegraphics[width=0.44\textwidth]{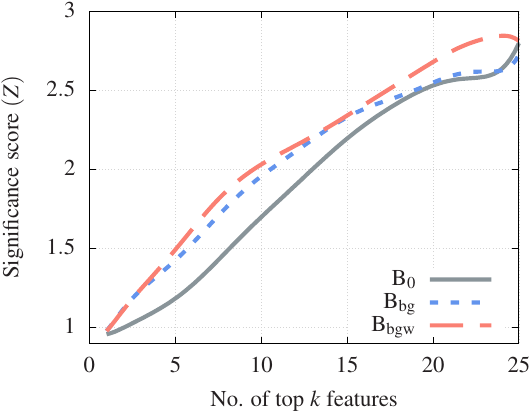}}\hfill
    \caption{Fig (a) show an increase in accuracy as we include more of the top-$k$ features for each of the baseline. Fig (b) shows an increase in the signal sensitivity ($Z$) as we include more of the top-$k$ feature for each of the baselines. We see that the averaged baselines consistently outperform the blank baseline for all values until $k\sim15$.}
    \label{fig:metrics}
    \end{center}
% \vskip -0.8em
\end{figure*}

\section{Experimental Setup}

\textbf{Dataset: }
We pick a typical particle physics search for a new particle to generate the dataset. We consider the production of two vectorlike quarks (VLQs) at the Large Hadron Collider (LHC). The task is to isolate the events from this process from the background processes with similar final states. The LHC is actively looking for the VLQs. They are hypothetical heavy quarks that appear in many beyond the Standard Model (SM) theories and are generally heavier than the heaviest known elementary particle, the top quark ($t$).

We consider the $p p \to B B$ process (the collision of two protons producing a pair of vectorlike $B$ quarks) as the signal~\cite{Bhardwaj:2022nko}. These heavy quarks decay soon after they are produced; we pick a signal where one $B$ decays to a bottom quark ($b$) and another hypothetical spinless state $\Phi$ (which further decays to two $b$ quarks), the other to a $t$ quark and a $W$ boson~\cite{Bardhan:2022sif}. 
We take the mass of the hypothetical $B$ quark to be $1.5$ TeV (which respects the current exclusion limits on such quarks) and the mass of the new scalar, $M_\Phi = 0.4$ TeV. All SM processes that produce the same final state form the background of our signal. For our purpose, we simulate ten different background processes, of which the monoleptonic $tt$ and $W+2j$ processes form the largest backgrounds. 
All the kinematic features of these objects (i.e., transverse momenta, pseudorapidity, etc.) and other event-level variables (e.g., the scalar sum of transverse momenta, $H_T$) form the total set of input variables (see Appendix). We generate roughly $1$M samples, out of which we use $20$\% of the data for validation and keep $30$\%  for testing. The remaining $50$\% is used to train the classifier. 
\medskip

\noindent \textbf{Model and training: }
We train an unbiased classifier on the classification task of signal vs background. The classifier comprises $2$ linear layers of hidden dim size 128, each followed by a \texttt{Swish} activation. \texttt{BatchNorm}, \texttt{Dropout} ($p=0.2$) and \texttt{Weight Decay} ($\lambda = 10^{-4}$) have been used to regularise the model. These hyperparameters were obtained by performing a grid search over a large set of values. The network is trained until convergence, and the model with the best loss on the validation dataset is chosen.
\medskip

\noindent \textbf{Implementing baselines:}
The blank baseline is constructed as a zero vector of the same dimension as the input. 
For the background average baseline, we sample equally from each background process, taking roughly 20 samples from each class, giving us a total of 200 baselines. The weighted baseline is then obtained by performing a weighted averaging of the attributions with weights as the cross-section. IG for each baseline is calculated for $5000$ signal inputs and averaged to give a signal class-level feature attribution. 
\medskip

\noindent \textbf{Measuring performance for the top-$k$ features: }
To evaluate the performance of the baselines, we propose to retrain the network using the top-$k$ important features reported by performing IG on both the baselines and evaluate its performance (for a particular top-$k$) in terms of accuracy and discovery sensitivity. We expect that for the baseline with the better feature ordering we would see rapid performance as we iteratively increase the top-$k$, with both baselines eventually plateauing at roughly the same performance a large value of $k$. 

\section{Results}

Fig.~\ref{fig:top5} shows the top $5$ features with the highest attributions for each of the baselines. Interestingly, $H_T$ has the highest feature attribution irrespective of the baseline, and this may be because the signal has distinctly higher $H_T$ than the backgrounds. This is because the mass scales of signal events are higher than that of the backgrounds. Similarly, we expect high attributions for features like $p_{T_\ell}$ (momentum of the reconstructed lepton) and $\slashed{E}_T$ (missing energy in the event) as we expected high values for these features in signal events since the final state objects are significantly boosted. This is missing in the top attributions of the blank baseline (See Appendix). We also note that, interestingly, the feature attributions for the invariant masses of the reconstructed jets ($m_{j1}, m_{j2}$, etc.) rank high (in the top 5) in both the $\text{B}_0$ and $\text{B}_{\text{bg}}$ baselines. These features (i.e. their distributions) do not provide any significant separability between signal and background events. For the $\text{B}_{\text{bgw}}$ baseline, it ranks much lower.

From Fig~\ref{fig:metrics}, we observe that the accuracy performance obtained from the top-$k$ features for both the unweighted and weighted averaged background baseline consistently outperforms the blank baseline. We see a similar trend for the $Z$ score. The performance rises until it plateaus for all the baseline around $k=12$, implying that the top-$12$ features are good enough to distinguish the signal from the background. We want to highlight the exact value of $k$ for which the performance would plateau would depend on the particular task at hand. We see only a small difference in performance between the weighted and unweighted averaged background baselines, with the weighted one performing slightly better. This warrants a more in-depth study.

We also ran an ablation test to analyze the feature attributions, but we found that it was difficult to obtain a clear pattern. This was mostly due to the large number of features used, most of which might contained overlapping information.

\section{Conclusion}

In this paper, we demonstrated how an interpretability technique like IGs can help us understand the workings of a DNN model. In a scenario where ``zero-information'' a.k.a. blank baselines are not helpful in interpreting model predictions through IGs meaningfully, we showed that carefully designed alternate baselines could provide insights. We illustrated this with a typical classification task in particle physics where one wants to separate new physics collision events from the SM background events. We showed that one can probe the model predictions with a baseline derived by averaging IGs over a set of background samples and identify what makes the signal events different from the background ones for the model. We showed that the proposed baseline strategy provided more reasonable feature attributions and found features with better discrimination ability compared to a blank baseline. This can be very useful for model explanation and verification, robust feature selection, and possibly even understanding the underlying physics processes involved. 

%In this paper, we demonstrated how an interpretability technique like IGs can be applied to understanding the workings of a DNN model. In a scenario where ``zero-information'' a.k.a. blank baselines, cannot be used to interpret model predictions through IGs meaningfully, we show that thoughtfully designed alternate baselines can provide insights into the model predictions. We illustrated this by picking a typical classification task in the particle physics domain where we want to separate new physics scattering events from the SM background events. We showed that one can probe the model predictions with a baseline derived by averaging IGs over a set of background samples to identify what makes the signal events different from the background ones for the model. We showed that the proposed baseline strategy not only provided more reasonable feature attributions but also found features that show better discriminative ability. This can be very useful for model explanation and verification, robust feature selection, and possibly even understanding the underlying physics processes involved. 

\medskip

\noindent\textbf{Limitations and scope  }

\noindent In future work, one can extend the work for other kinds of tasks relevant to machine learning in HEP, such as jet classification, as well as other types of models, such as GNNs. We believe that interpretability should become a primary priority for machine learning endeavours. In the future, one could apply interpretability methods to improve their understanding of physics directly from large models trained on vast amounts of data. 

\bibliography{aaai25}

\newpage
\clearpage

\onecolumn
\appendix

\section{Feature Attribution List}
\label{app:features}
We plot below the feature attribution list for the top-$20$ variables for all the baselines.

\begin{figure}[htp]
    \centering
    \includegraphics[width=0.9\linewidth]{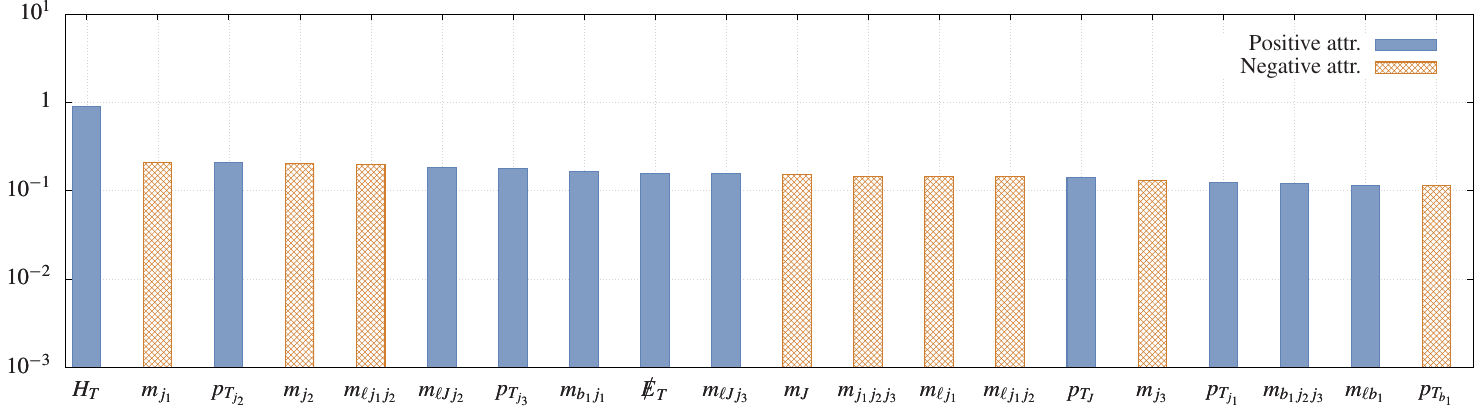}
    \caption{Top-$20$ features ranked by attribution for baseline B$_0$}
    \label{fig:attr-list-b0}
\end{figure}

\begin{figure}[htp]
    \centering
    \includegraphics[width=0.9\linewidth]{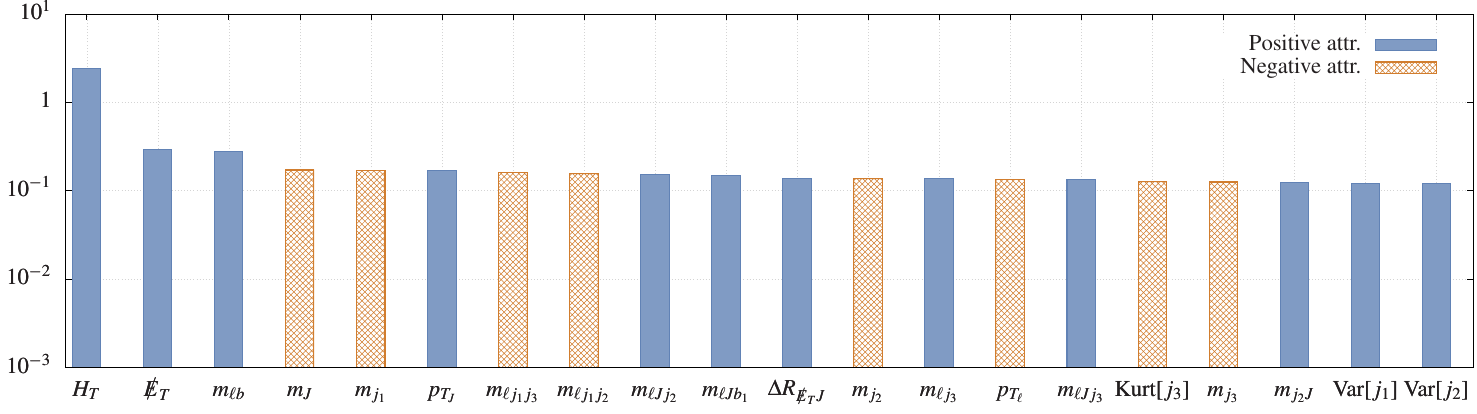}
    \caption{Top-$20$ features ranked by attribution for baseline B$_{\text{bg}}$}
    \label{fig:attr-list-bg}
\end{figure}

\begin{figure}[htp]
    \centering
    \includegraphics[width=0.89\linewidth]{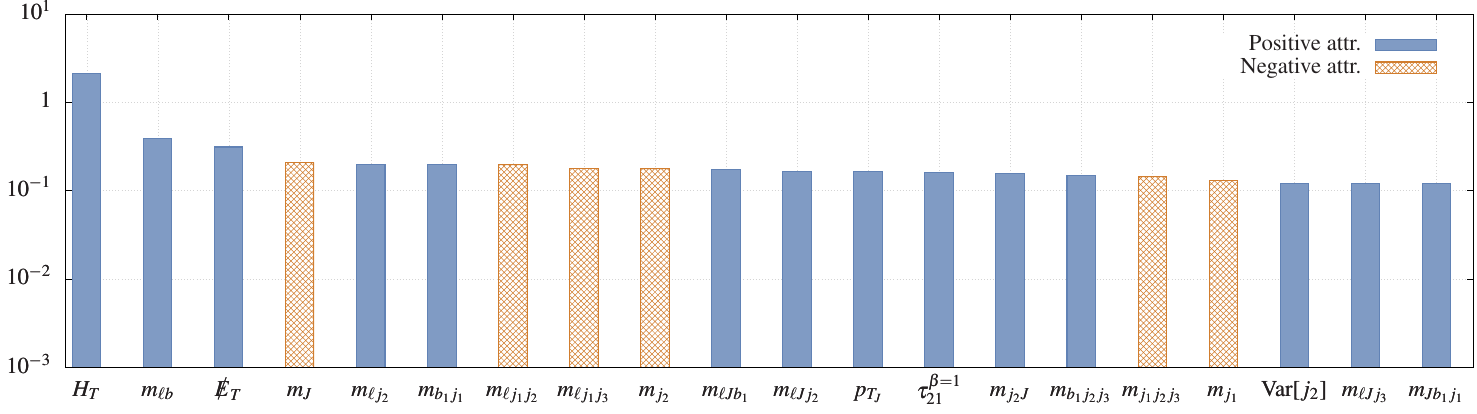}
    \caption{Top-$20$ features ranked by attribution for baseline B$_{\text{bgw}}$}
    \label{fig:attr-list-bgw}
\end{figure}

\section{Kinematic and Topological Features}
\label{app:input_var}
The network is trained on different kinematic distributions of the signal and background events. Each selected event has some well-defined objects---the high-$p_T$ lepton, the three AK-4 jets, the $b$-tagged jet, the fatjet, and missing $E^{\text{miss}}_T$  (since the lepton in the signal comes from the decay of a $W$-boson). We feed the network the following kinematic properties of these objects:

\begin{enumerate}
    \item \emph{Basic variables:} For each identified object, we consider the transverse momentum $(p_T)$. The scalar $H_T$ of the event and missing energy is also considered. The set of kinematic variables chosen is $\left\{H_T, |E^{\text{miss}}_T|, p_{T_\ell}, p_{T_{j1}}, p_{T_{j2}}, p_{T_{j3}}, p_{T_b}, p_{T_J}\right\}$.
    
    \item \emph{Jet-substructure variables:} For the fatjet, we calculate the $n$-subjettiness ratios $(n=1,2,3)$ for multiple $\beta$ values $(\beta = 1.0, 2.0)$ to take the prongness of $J$ into account. The set used is $\left\{ \tau_{21}^{\beta=1},\tau_{21}^{\beta=2}, \tau_{32}^{\beta=1}, \tau_{32}^{\beta=2}\right\}$.
    
    \item \emph{Distance in the $\eta-\phi$ plane:} We calculate the separation between two objects as $\Delta R_{ij} = \sqrt{\Delta \Phi_{ij}^2 + \Delta \eta_{ij}^2}$. We choose all possible pairs from the reconstructed objects and calculate the distance between them.
    
    \item \emph{Invariant masses of objects and their combinations:} We consider the masses of hadronic objects and the invariant masses of combinations of $2$ or $3$ objects. The set of (invariant) mass variables is  $\left\{m_{i'}, m_{ij}, m_{ijk} \right\}$, where $i'$ denotes an reconstructed hadronic object and each of $i$,$j$, and $k$ represent any reconstructed object.
    
    \item \emph{Girth/width of hadronic objects:} The girth (width) of a hadronic object is the $p_T$-weighted average distance of the constituents of the jet from the jet axis. We also consider the higher-order central moments---variance, skewness and kurtosis of the distribution. The set of girth (width) related variables considered is $\left\{ g_{i'}, \text{Skew}[i'], \text{Kurt}[i'], \text{Var}[i']\right\}$, where $i'$ denotes a reconstructed hadronic object and Skew, Kurt, Var stand for the skewness, kurtosis, and the variance of the $p_T$-weighted distribution of the constituents of the hadronic object.
\end{enumerate}

\end{document}